# Trustable and Automated Machine Learning Running with Blockchain and Its Applications

Tao Wang
SAS Institute Inc.
Cary, NC, USA
t.wang@sas.com

Xinmin Wu
SAS Institute Inc.
Cary, NC, USA
Xinmin.Wu@sas.com

Taiping He
SAS Institute Inc.
Cary, NC, USA
Taiping.He@sas.com

## ABSTRACT

Machine learning algorithms learn from data and use data from databases that are mutable; therefore, the data and the results of machine learning cannot be fully trusted. Also, the machine learning process is often difficult to automate. A unified analytical framework for trustable machine learning has been presented in the literature. It proposed building a trustable machine learning system by using blockchain technology, which can store data in a permanent and immutable way. In addition, smart contracts on blockchain are used to automate the machine learning process. In the proposed framework, a core machine learning algorithm can have three implementations: server layer implementation, streaming layer implementation, and smart contract implementation. However, there are still open questions. First, the streaming layer usually deploys on edge devices and therefore has limited memory and computing power. How can we run machine learning on the streaming layer? Second, most data that are stored on blockchain are financial transactions, for which fraud detection is often needed. However, in some applications, training data are hard to obtain. Can we build good machine learning models to do fraud detection with limited training data? These questions motivated this paper; which makes two contributions. First, it proposes training a machine learning model on the server layer and saving the model with a special binary data format. Then, the streaming layer can take this blob of binary data as input and score incoming data online. The blob of binary data is very compact and can be deployed on edge devices. Second, the paper presents a new method of synthetic data generation that can enrich the training data set. Experiments show that this synthetic data generation is very effective in applications such as fraud detection in financial data.

## 1 Introduction and literature review

Many machine learning algorithms suffer from two common problems: trustability and lack of automation. First, it can be difficult to trust the results from a machine learning algorithm because machine learning algorithms use data from databases that are mutable. System administrators and illegal hackers can modify the data source, and this will eventually change the results, with or without notification. Second, it can be difficult to automate the machine learning pipeline. Currently, the machine learning pipeline is mostly controlled and monitored by human beings. Sometimes this process might begin or end at suboptimal times because of human involvement and the imperfect nature of human beings.

A unified analytical framework [24] for trustable machine learning has been presented in the literature. It proposed building a trustable machine learning system by using blockchain technology. A blockchain [1] is a continuously growing, single-linked list of immutable records (blocks) that are often secured using cryptography. Blockchain was first invented by Satoshi Nakamoto [1] in 2008 as a public financial transaction ledger for use in the cryptocurrency Bitcoin. Blockchain solved the Byzantine Generals Problem [3] by using a peer-to-peer (P2P) system without going through a trusted financial institution. The double-spending problem [4, 5] was also solved in a pure P2P decentralized network without any financial institution involved. This P2P network timestamps transactions by hashing them using SHA-256 [23] into an ongoing blockchain of hash-based proof-of-work (PoW), forming a record (block) that cannot be changed without redoing the PoW (also known as mining), which involves a substantial amount of computing power. The longest blockchain with the highest combined difficulties serves not only as proof of the sequence of transactions witnessed but also as proof that it came from the largest pool of computing power. With more and more computers added to the blockchain every day, it is increasingly difficult to hack the blockchain system unless the hacker overpowers the rest of the world, which is almost impossible in practice. So, people believe the data that are saved in blockchain are immutable and therefore can be fully trusted. The paper [24] proposed that machine learning algorithms should use the immutable data provided by blockchain to solve the trustability problem.

While Bitcoin is widely considered to be blockchain 1.0, Ethereum [2] is often considered to be blockchain 2.0, because it used blockchain not only as the foundation for cryptocurrency but also for decentralized applications (DApps) and decentralized autonomous organizations (DAOs). The Ethereum network provided a blockchain with a built-in, fully fledged, Turing-complete [6] programming language that can be used to implement so-called "smart contracts". Smart contracts are essentially automated processes that can be used to encode arbitrary state transition functions, allowing one to create and run complicated systems (such as Facebook and Twitter, theoretically) on top of the Ethereum blockchain. The Ethereum blockchain opened a door to the largest development effort witnessed so far in the world of blockchain. Even some "traditional" companies like Kodak [9] are tapping into the Ethereum blockchain technology. The paper [24] proposed to use the smart contract as the automation engine to solve the automation problem. In the proposed framework [24], a core machine learning algorithm can have three implementations: server layer implementation, streaming layer implementation, and smart contract implementation. A server layer implementation is the implementation of the machine learning algorithm after code refactoring so that it can run on top of a server layer, which is a cloud-based computing environment. Usually, a server layer can have two running modes: symmetric multiprocessing (SMP) and massively parallel processing (MPP). A streaming layer implementation is the implementation of the machine learning algorithm after code refactoring so that it can run on top of the streaming layer. Usually, a streaming layer is a computing environment that runs tasks in sliding windows and discards old data after use. A major difference between the streaming layer and the server layer is that the streaming layer handles data events on the fly (dynamically) and the server layer handles all the data at rest (statically). Another difference is that the streaming layer often removes old data to make room for new data while the server layer usually does not remove old data. A smart contract implementation is the implementation of the machine learning algorithm after code refactoring so that it can run on top of the smart contract layer. As mentioned previously, a smart contract is an automated process. Once a machine learning algorithm is implemented as a smart contract and running on a blockchain, the automation problem can be largely solved or alleviated. This is because when the predefined conditions are satisfied, the automated process will be triggered and will run on the blockchain.

However, a few questions remain open. First, the streaming layer usually deploys on an edge device that has limited memory and computing power. How can we train or score a machine learning model with limited memory and computing power? Second, nowadays, most data stored on blockchain are financial transactions that often require fraud detection. For example, the J.P. Morgan Interbank Information Network (IIN) [25] requires fraud detection before a financial transaction enters the blockchain. However, in such applications, sometimes it is hard to obtain the training data for training a good fraud detection model. Can we build a good fraud detection model with limited training data? This paper aims to answer both questions. First, it proposes training a machine learning model on the server layer and saving the trained model with a special binary data format. Then, the streaming layer can take this blob of binary data as input and score incoming new data in an online fashion. The blob of binary data is very compact in size and can be deployed on edge devices. Second, it presents a new method of synthetic data generation that can enrich the limited training data set. Experiments show that this synthetic data generation is very effective in applications such as fraud detection in financial domains.

The rest of this paper is organized as follows. Section 2 reviews the prior art. Section 3 introduces the special binary data format and describes how to save machine learning models with this format. Section 4 presents the synthetic data generation algorithm with experimental results. Section 5 concludes the paper and points out some future research directions.

## 2 Review of prior art

In the literature, a paper [24] proposed the use of blockchain technology to solve the trustability problem and the use of smart contract to solve the lack of automation problem for machine learning.

The paper [24] classified machine learning algorithms into four categories: supervised machine learning, unsupervised machine learning, semi-supervised machine learning, and all others. Supervised machine learning requires the target (dependent) variable Y to be labeled in the training data set so that a model can be built to predict the label of unseen data, and it often discards training observations that have unlabeled targets. Unsupervised machine learning does not require the target (dependent) variable Y to be labeled in the training data, and its goal is not to predict the label but rather to infer a function of or to summarize the unlabeled training data. Semi-supervised machine learning requires the target (dependent) variable Y to be labeled in only a small part of the training data, and it is often used as a preprocessing stage for supervised machine learning when labeling all the training data is impossible or too expensive. And there are many other machine learning algorithms, such as reinforcement learning, association rule mining [11], adversarial learning, and so on. Although machine learning algorithms vary a lot, most of them have the following seven steps, assuming the data are already prepared and clean:

1) Model initialization. In this step, the machine learning algorithm initializes the model by checking required licenses and system resources (such as GPU or other devices) and by allocating memory for the data structures.

2) Model training. In this step, the machine learning algorithm trains a model by using the training data set. This is usually one of the most important steps. The trained model should be a good representation of the training data, excluding the noise.

3) Model validation. This step is optional, but it is crucial to many machine learning algorithms to prevent overfitting. Overfitting happens when the model represents the training data very well but also memorizes the noise. In this step, the machine learning algorithm validates the trained model on the validation (holdout) data set to test if the trained model can make good predictions to unseen data.

4) Model scoring (testing). This step is optional. Some unsupervised machine learning algorithms have no model scoring step. For supervised machine learning, model scoring is very important and often required. For prediction and regression tasks, the basic idea of scoring is to use the trained model to make predictions with new or unseen data.

5) Model evaluation (assessment). This step is optional. The basic idea of assessment is to evaluate the quality of the trained model according to certain metrics. Machine learning algorithms can be compared with each other through model evaluation.

6) Model serialization (persistence). This step is optional. To ensure that a trained model can be used to score unseen data, the model needs to be serialized and saved for future use.

7) Model clean-up. This is often the last step in many machine learning algorithms. In this step, the algorithm frees the allocated memory and releases obtained system resources.

The paper [24] used these seven steps to demonstrate the unified analytical framework for trustable machine learning and automation that runs on blockchain. The framework can be visualized with Fig. 1.

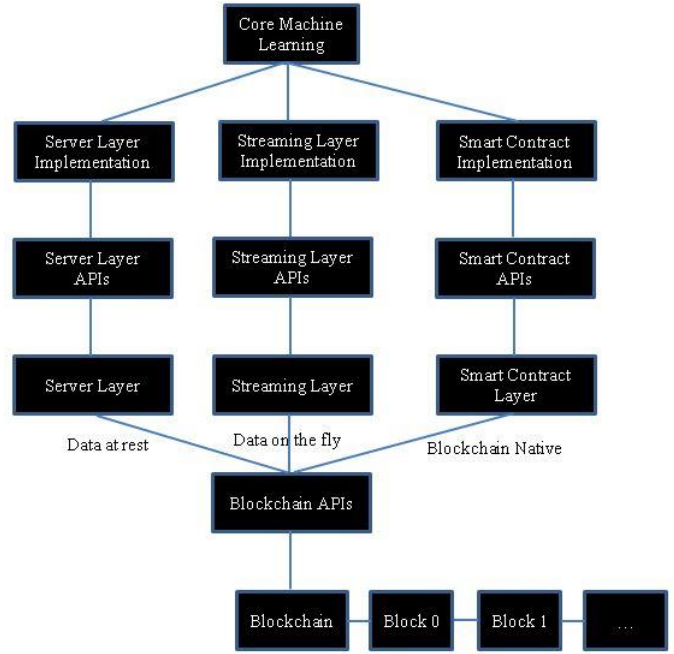

Fig. 1. An analytical framework for trustable machine learning.

The following paragraphs explain each component of Fig. 1 in more detail.

1. Core machine learning is the implementation of machine learning algorithm in its native form. It often includes the seven components described earlier in this paper: model initialization, model training, model validation, model scoring, model evaluation, model serialization, and model cleanup.

2. Server layer implementation is the implementation of the machine learning algorithm after code refactoring so that it can run on top of the server layer which is a cloud-based computing environment. The server layer often has two running modes: symmetric multiprocessing (SMP) and massively parallel processing (MPP).

3. Streaming layer implementation is the implementation of the machine learning algorithm after code refactoring so that it can run on top of the streaming layer. A streaming layer is a computing environment that runs in sliding windows and discards old data after use.

4. Smart contract implementation is the implementation of the machine learning algorithm after code refactoring so that

it can run on top of the smart contract layer. As mentioned previously in this paper, a smart contract is just an automated process. Once a machine learning algorithm is implemented as a smart contract and running on blockchain in a native way, the automation problem can be solved or largely alleviated. This is possible because when the predefined conditions are satisfied, the smart contract will be triggered and will run on the blockchain. Once running, it cannot be canceled unilaterally.

5. Server layer APIs are the APIs provided by the server layer. Currently, most (if not all) server layer offerings come with SDK, which is a complete set of APIs that enable one to create applications to run on the server layer.

6. Streaming layer APIs are the APIs provided by the underneath streaming layer. Currently, most (if not all) streaming layer offerings come with SDK, which is a complete set of APIs that enable one to create applications to run on the streaming layer.

7. Smart contract APIs are the APIs provided by the underlying smart contract layer. Currently, most (if not all) smart contract layer offerings come with SDK, which is a complete set of APIs that enable one to create applications to run on the smart contract layer.

8. Server layer is a cloud-based computing environment that can train or score machine learning models.

9. Streaming layer is a computing environment that can train or score machine learning models in sliding windows.

10. Smart contract layer is a computing environment that can train or score machine learning models on blockchain as a native application.

11. Blockchain APIs are the APIs provided by the underlying blockchain. The server layer can obtain aggregated data at rest (in one shot or multiple shots) from the blockchain via blockchain APIs. The streaming layer can obtain live data on the fly from the blockchain via blockchain APIs. The smart contract layer can obtain data from the blockchain via blockchain APIs in a native way.

12. Blockchain is a continuously growing, single-linked list of immutable records or blocks.

The seven common steps of machine learning can have server layer implementation, streaming layer implementation, and smart contract implementation. The details can be found in the paper [24].

## 3 Saving machine learning models with ASTORE

The unified analytical framework that is proposed in [24] uses blockchain to address the two common problems of machine learning: trustability and lack of automation. However, a few problems remain to be solved. First, the streaming layer usually deploys on an edge device, usually has very limited memory and poor computing power. How can we train or score a machine learning model on such devices? This section aims to solve this problem.

In this section, we propose training a machine learning model on the server layer and saving the trained model with a special binary data format: ASTORE [21]. Then, the streaming layer can take this blob of ASTORE data as input and score incoming new data as they come in. The ASTORE data is very compact in size and can be deployed on edge devices.

The ASTORE (analytic store) is a binary format for machine learning models. It was designed to be unique and immutable. Therefore, we found it very attractive for tasks that are related to blockchain.

An ASTORE of a machine learning model is essentially a serializable binary object. It contains a unique store key, which is universal and is secured using cryptography. ASOTRE saves the machine model states, and any information needed to recover a model, into a platform-independent binary blob. This binary blob can be stored in a local file, a blob table in the cloud, a blob in the databases, or in other formats. An ASTORE blob contains model information, the shared library, the entry function that is required to run this ASTORE, lists of input and output variables, and so on. All this information is packed and serialized when the blob is created, and it is unpacked and deserialized when the blob is loaded into memory. Once an ASTORE is created, it can be transferred and used on any platform. Therefore, it is very flexible and can score a data set in different environments.

Here is a list of detailed information in an ASTORE:

- store key
- model name and description
- algorithm information
- timestamp of model creation
- training parameters
- score functions and rules
- input variables, data types, and data formats
- output variables, data types, and data formats

One of the key features of the ASTORE format is that its store key is generated and secured using cryptography. The store key contains a string of characters such as: 2580E6ABBCEE8B9C05689CDD952C60554A76A02E. This store key ensures that each ASTORE data file is unique and immutable. If one bit of data is changed, the store key is changed too. This makes ASTORE compatible with blockchain.

Since 1997, several machine learning model formats, such as PMML [18], PFA [19], and ONNX [20], have been introduced. PMML [18] uses Extensible Markup Language (XML) to represent a machine learning model; that is, PMML models are described in XML schema. One or multiple models can be in one PMML document. PMML is broadly used in the machine learning community and is supported by commercial enterprise software such as SAS, SPSS, SAP, KNIME, TIBCO, Microsoft, and Teradata. It is also supported by open source tools such as Apache Spark, Weka, and R. PFA [19] is a mini-language for numerical computations. A PFA document is described by JSON or YAML scripts. It can be just a simple data transformation, or it can be a combination of several complicated machine learning models. Unlike PMML, which focuses simply on model description, PFA focuses on the scoring procedures. PFA is platform independent: the scoring engine in a PFA-enabled system is guaranteed to generate the same output from the same input. PFA supports a wide range of existing models, and it also supports new models through action definitions and flow control operations. PMML and PFA specifications are maintained by the Data Mining Group (DMG) [45]. ONNX [20] is another format that is used to store the machine learning models. It supports meta-data description through key-value pairs. It also supports various operations that are based on the data flows, but it focuses mainly on deep learning models and it is heavily Python-dependent. ONNX is sponsored by many large commercial companies such as Amazon, Facebook, Microsoft, SAS, and so on. For the latest development, see [46]. Since PMML, PFA, and ONNX models are all plain-text files, they can be reverse-engineered. This is a big security concern for commercial use, especially for financial applications.

Because ASTORE is a binary format, it is almost impossible to reverse-engineer it. Therefore, it is a better choice for the streaming layer and the smart contract layer. We implemented ASTORE utility functions, which can describe the detailed information inside an ASTORE model and perform scoring with the input data set. Also, we optimized the ASTORE blob in a multithreaded environment, which makes the scoring process very efficient in SMP mode, MPP mode, and in streaming environment. An ASTORE blob is platform-independent, so it can be used in different operating systems: on a local machine, in a database, in the cloud, or even in GPUs. The ASTORE format supports all types of supervised machine learning algorithms, from simple linear regression to deep neural network.

In Pseudocode Snippet 1, the function "describe" provides

**Pseudocode Snippet 1: Deploy ASTORE Model**

```
function astore_score()
    input:   astore_model
             input_data_set
    output:  output_data_set

    status = OK;
    status = describe()
        input:   astore_model
        output:  input_variable_list
                 output_variable_list
    if (status != OK) goto Finish;

    status = input_data_set_validation()
        input:   input_data_set
                 input_variable_list
    if (status != OK) goto Finish;

    status=setup_output_data_set()
        input:   output_variable_list
        output:  output_data_set
    if (status != OK) goto Finish;

    for each input_data_value in input_data_set do {
        status = score()
            input:   input_data_value
            output:  output_data_value
        if (status != OK) goto Finish;

        status = insert_output_data()
            input:   output_data_value
            output:  output_data_set
        if (status != OK) goto Finish;
    }

    status = write_output_data_set()
        input:  output_data_set
    if (status != OK) goto Finish;

Finish:
    return status;
```

the basic ASTORE model information, including input variables and output variables. The function "input_data_set_validation" checks the variable attributes of the input data set to make sure the input data are valid. The function "setup_output_data_set" creates a data set with valid output variables, which are obtained from the "describe" function. The main function "score" calculates the score result for the input data. The function "insert_output_data" adds a single score result to the output data set, and the function "write_output_data_set" delivers the entire output data.

As mentioned previously, the first problem we aim to address is that the streaming layer usually deploys on an edge device that has very limited memory and poor computing power. Sometimes, it is unwise to train a complicated machine learning model in such environment. We propose training the machine learning model on the server layer, which is often running in a powerful cloud environment. The trained model is then saved by using the ASTORE format. Once the ASTORE model is ready, the streaming layer can take it as input and score incoming new data as they arrive. The ASTORE data is very compact in size and can be deployed on edge devices. The first problem is solved with ASTORE.

## 4 Synthetic data generation using capsules

Nowadays, most data that are stored on blockchain are financial transactions that often require fraud detection. For example, the J.P. Morgan Interbank Information Network (IIN) [25] requires a fraud detection process before a financial transaction enters the blockchain. However, in such applications, sometimes the training data is difficult to obtain to train a good fraud detection model. Can we build good fraud detection model with limited training data? In this section, we present a new method of synthetic data generation that can enrich the training data set. Experimental results show that this synthetic data generation is very effective for applications such as fraud detection in financial domains. Some may argue that synthetic data generation is a relatively separate topic. It is included in this paper because it enables one to train a good model with limited training data which is exactly the situation we faced here.

Synthetic data generation has played an important role [26] in the areas of data science and machine learning. It is widely deployed in industries such as financial services, and healthcare, where targeted events are rare, or it is hard to collect enough training data. In addition, synthetic data can help meet customized needs or simulate conditions that are simply not available in a real environment because of legal or privacy concerns. Synthetic data can enrich the data source to build better models. Also, once the synthetic data generation engine is ready, producing a lot of data is fast and cheap. Synthetic data generation algorithms can produce perfectly accurate labels for target variables, which is hard to achieve in real data. Synthetic data generation has been widely used in many applications including training self-driving cars [27], point tracking [28], and transfer learning [29].

Some research has been done on measuring characteristics in data and how they affect rare-event detection systems [47,48]. The anticipated impact of synthetic "big" data on learning analytics (LA) infrastructures was presented in [49], with a particular focus on data governance, the acceleration of service development, and the benchmarking of predictive models. The authors of [49] argue that the application of synthetic data not only will accelerate the creation of complex and layered learning analytics infrastructure but will also help to address the ethical and privacy risks that are involved during service development. The paper [50] generates "random" synthetic data with different degrees of regularity and shows that it affects the false alarm rate drastically. In [51], an experiment aimed at generating synthetic test data for fraud detection in an IP-based video-on-demand service is studied. The synthetic data generation in [52] is a simulation-based development that ensures that important statistical properties of the authentic data are preserved by using authentic normal data and fraud as a seed for generating synthetic data. The data generation has five steps: data collection, data analysis, key parameter of fraud detection identification, user model, system model. As we can see, building those simulators is fairly complex and also might involve human judgement. Our method of synthetic data generation is different. It can automatically identify the key features from fraud and non-fraud from real data, and then re-create data based on the learned capsule embeddings. The process can provide a lot of synthetic data in very short time. In addition, existing popular synthetic data generation method such as PSDG and SDDL [55–56] are primarily based on fitting statistical distributions, but our capsule network method focuses on learning embeddings.

In this paper, we use Capsule Network (CapsNet) [30] to generate synthetic data for training the fraud detection model. Capsule Network is a new addition to the deep learning family. It has a few unique features and advantages [30] over convolutional neural network (CNN). Despite of CNN's good performance and success in various tasks such as image segmentation, imagine classification, and object detection, it is well-known to have some drawbacks such as the lack of data reconstruction capability [31]. In the context of this paper, synthetic data generation is essentially data reconstruction. Fortunately, Capsule Network can be used for data reconstruction and synthetic data generation.

A capsule is a group of neurons whose activity vector represents the instantiation parameters of a specific type of entity such as an object or a part of an object [30]. A capsule extends a neuron from a scalar to a vector. So, the output of a capsule is a function of an input vector. A squash function was used as the activation [30], where $s_j$ is the vector input of capsule $j$ and $v_j$ is the output of capsule $j$:

$$v_j = \frac{\|s_j\|^2}{1+\|s_j\|^2} \frac{s_j}{\|s_j\|} \qquad (1)$$

In addition, dynamic routing in [30] was proposed to determine how a lower-level capsule sends its output to higher level-capsules according to the agreements between the outputs of lower-level capsule and the activity vectors of higher-level capsules. The vector input to a capsule $j$, $s_j$ is a weighted sum over all prediction vectors $\widehat{u_{j|i}}$ from the capsules in the previous layer and is produced by multiplying the output $u_i$ of a capsule in the previous layer by a weight matrix $W_{ij}$:

$$s_j = \sum_i c_{ij} \widehat{u_{j|i}}, \quad \widehat{u_{j|i}} = W_{ij} u_i \qquad (2)$$

where $c_{ij}$ are coupling coefficients that are determined by the dynamic routing process:

$$c_{ij} = \frac{\exp(b_{ij})}{\sum_k \exp(b_{ik})} \qquad (3)$$

$$\sum_j c_{ij} = 1 \qquad (4)$$

Data reconstruction with labels as a regularization method is introduced in [30] to improve image classification of the MNIST data set [32]. Capsule Network masks out all but the activity vector that is associated with the correct output capsule, and only that activity vector is used to reconstruct the input data. Essentially, each capsule learns the embedding that is related to its label. Therefore, data that are reconstructed in this way can be used to improve classification. Based on this fact, we chose to use Capsule Network to do synthetic data generation to improve fraud detection.

The architecture of synthetic data generation based on Capsule Network is shown in Fig. 2.

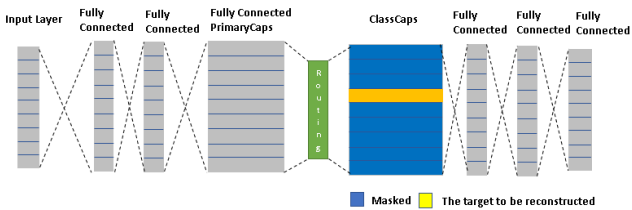

Fig. 2: Synthetic data generation based on Capsule Network

The synthetic data generation architecture that is based on Capsule Network is very flexible. It can connect with all the popular deep learning networks such as VGG [33], ResNet [34], or DenseNet [35]. It can handle not only image, speech, and text data, but also tabular data. In this paper, we use this architecture to conduct synthetic data generation to improve fraud detection with data coming from blockchain and saved in memory as tabular data. This architecture is similar to an autoencoder [36] – the input layer is the same with the output layer. During training, this architecture builds an embedding of the input data so that the input data is memorized. During scoring, we change the parameters of the trained model to generate synthetic data.

Although this proposed architecture is similar to an autoencoder, there are a few differences. These differences are also our contributions. First, this architecture can support various data types such as image, text, voice, and tabular data. Secondly, the data mask in the ClassCaps layer allows the capsule embedding to learn only the feature from its desired label. This is not only the biggest advantage of doing reconstruction using Capsule Network but also the biggest difference from an autoencoder. Last but not least, this architecture learns embedding from the reconstruction process. The embedding can be used as inputs to other tasks such as classification, clustering, regression, and so on. Compared to single task learning, this multi-task learning [37] framework can improve those tasks because reconstruction serves as a regularization method [30].

In this section, we will apply this architecture to perform synthetic data generation for tabular data. In this architecture, each element in the capsule learns a subset of input features. Therefore, we can tweak some of the elements within a capsule to generate more data that are different from reconstruction outputs but retain much of the learned features. More importantly, the capsule embedding that is learned from reconstruction can serve as input for various learning processes with much less dimensionality downstream.

To illustrate that this synthetic data generation architecture can create more training data to improve financial fraud detection, we tested our algorithm on the Credit Card Fraud Detection data set [38] from Kaggle. The data set contains credit card transactions by European cardholders in September 2013. It presents transactions that occurred in two days, with 492 fraudulent transactions out of 284,807 transactions. The data set is highly unbalanced: the positive class (fraudulent transactions) account for 0.172% of all transactions. The data set has 29 numerical variables and a binary target (fraud or non-fraud). In the data preparation step, we applied stratified sampling with a 50% ratio to split the whole data set into a training set and a test set. The training set is used for training the fraud detection model, and the test set is used to assess the model performance. For training, we applied an oversampling technique to ensure that the rare events and non-events are balanced.

Different machine learning algorithms such as SVM, random forest, and neural network have been applied to do anomaly detection in supervised learning settings where data is usually highly unbalanced. Anomaly detection with a learning-to-rank approach was introduced in [39]. The study found that gradient boosting has proved to be a powerful method on real-life data sets to address learning-to-rank problems because it optimizes in function space instead of parameter space and because it strengthens the impact of the rare event by a boosting strategy. In this paper, a gradient boosted tree [40] is used to train the model because of its excellent performance in fraud detection. It also serves as the baseline to measure enhancement of synthetic data generation. This architecture takes the data (after oversampling) as the input. The feature extraction layer is fully connected with 100 neurons. After the feature extraction layer, there is a reshape layer that has 10 capsules, with each capsule having 10 neurons. The classification layer has two capsules, which represent the binary target (fraud or non-fraud) for each target class. Each capsule in the classification layer contains 16 neurons. After the classification layer, there are three fully connected layers and the last layer with 29 neurons. We used the Adam SGD [41] algorithm with 250 epochs to train the model. After training, we scored the input data set to obtain the set of generated data from the layers. During scoring, we also introduced the same parameter ratio to tweak the capsule embedding in the model so that the generated data sets can present more diversity. Then, generated rare event data were added to the training set for modeling (denoted by Model 2), and we applied the exact same gradient boosted tree to do training. For a fair comparison, we also introduced a third model (denoted by Model 3), in which a set of rare events was generated by random number generation and added to the training set. For model assessment, we used the following three widely adopted metrics: precision recall curve [42], F1 curve [43], and ROC curve [44]. Since the event prior is only 0.172%, we compared performance over cutoffs only between 0 and 10%. The comparisons among the baseline model, Model 2, and Model 3 are shown in Figs. 3-5. We can see that overall the model that used synthetic data from Capsule Network performed better than the other two models in terms of balancing catching frauds and avoiding too many false alerts.

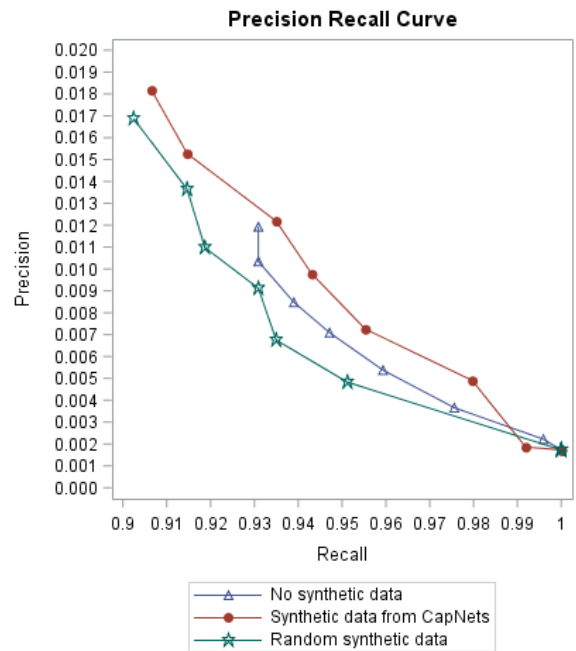

Fig. 3: Comparison with precision recall curve

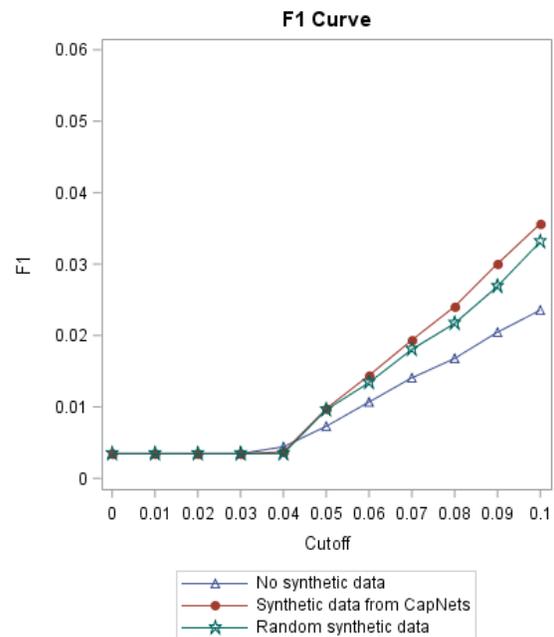

Fig. 4: Comparison with F1 curve

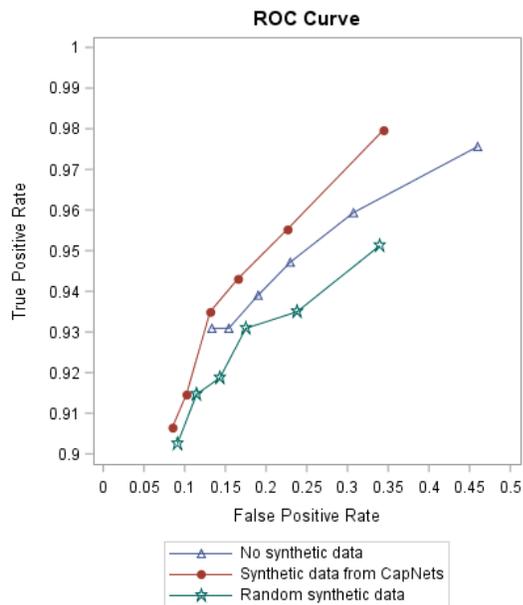

Fig. 5: Comparison with ROC curve

## 5 Conclusion and future research

In the literature, a paper [24] proposed the use of blockchain technology to solve the trustability problem and use smart contract to solve the lack of automation problem for machine learning. In the proposed framework, a core machine learning algorithm can have three implementations: server layer implementation, streaming layer implementation, and smart contract implementation. However, there are still open questions. First, the streaming layer usually deploys on edge devices that have limited memory and computing power. How can we run machine learning on these devices? Second, data stored on blockchain are often financial transactions for which fraud detection is usually required. However, in some applications, the training data are hard to obtain. Can we build good machine learning models to do fraud detection with limited training data? These two questions are the motivations of this paper which has two contributions. First, it proposes training a machine learning model on the server layer and saving the model with a special binary data format: ASTORE. Then, the streaming layer can take this blob of ASTORE data as input and score incoming data online. The ASTORE data are very compact and can be deployed on edge devices. Second, it presents a new algorithm of synthetic data generation, which can enrich the training data set. Experiments show that this synthetic data generation architecture is effective on applications such as fraud detection in financial data.

Blockchain faces some challenges such as how to make the blockchain infrastructure scalable and how to verify the computation on the geographically distributed system. Some We have seen some efforts such as using the IPFS (Inter-Planetary File System) [12] or IPDB (Inter-Planetary Database) [53]. However, according to the impossible triangle [54], a trade-off must be made among decentralization, consistency and scalability, and it is impossible to achieve them all at the same time. For machine learning applications, consistency and scalability are very important therefore some people may choose to relax the requirement for decentralization to certain degree. Please note that blockchains often run on machines equipped with GPU which can significantly enhance machine learning performance for many applications.

In the future, we plan to work on the smart contract implementation and determine whether the ASTORE data can be used as its input. We also like to enhance the ATORE structures to support feature transformation and provide more model description details. Because training machine learning models on blockchain in a native way can be inefficient and expensive, it makes more sense for the smart contract implementation to take the ASTORE model as its input and score new data coming from new blocks.